# Linear Regression with Limited Observation

Elad Hazan                                                                               ehazan@ie.technion.ac.il
Tomer Koren                                                                              tomerk@cs.technion.ac.il
Technion — Israel Institute of Technology, Technion City 32000, Haifa, Israel

## Abstract

We consider the most common variants of linear regression, including Ridge, Lasso and Support-vector regression, in a setting where the learner is allowed to observe only a fixed number of attributes of each example at training time. We present simple and efficient algorithms for these problems: for Lasso and Ridge regression they need the same total number of attributes (up to constants) as do full-information algorithms, for reaching a certain accuracy. For Support-vector regression, we require exponentially less attributes compared to the state of the art. By that, we resolve an open problem recently posed by Cesa-Bianchi et al. (2010).

Experiments show the theoretical bounds to be justified by superior performance compared to the state of the art.

## 1. Introduction

In regression analysis the statistician attempts to learn from examples the underlying variables affecting a given phenomenon. For example, in medical diagnosis a certain combination of conditions reflects whether a patient is afflicted with a certain disease.

In certain common regression cases various limitations are placed on the information available from the examples. In the medical example, not all parameters of a certain patient can be measured due to cost, time and patient reluctance.

In this paper we study the problem of regression in which only a small subset of the attributes per example can be observed. In this setting, we have access to *all* attributes and we are required to *choose* which of them to observe. Recently, Cesa-Bianchi et al. (2010)

studied this problem and asked the following interesting question: *can we efficiently learn the optimal regressor in the attribute efficient setting with the same total number of attributes as in the unrestricted regression setting?* In other words, the question amounts to whether the information limitation hinders our ability to learn efficiently at all. Ideally, one would hope that instead of observing all attributes of every example, one could compensate for fewer attributes by analyzing more examples, but retain the same overall sample and computational complexity.

Indeed, we answer this question on the affirmative for the main variants of regression: Ridge and Lasso. For support-vector regression we make significant advancement, reducing the parameter dependence by an exponential factor. Our results are summarized in the table below [1], which gives bounds for the number of examples needed to attain an error of $\varepsilon$, such that at most $k$ attributes [2] are viewable per example. We denote by $d$ the dimension of the attribute space.

| Regression | New bound | Prev. bound |
|:---:|:---:|:---:|
| Ridge | $O\left(\frac{d}{k\varepsilon^2}\right)$ | $O\left(\frac{d^2 \log \frac{d}{\varepsilon}}{k\varepsilon^2}\right)$ |
| Lasso | $O\left(\frac{d \log d}{k\varepsilon^2}\right)$ | $O\left(\frac{d^2 \log \frac{d}{\varepsilon}}{k\varepsilon^2}\right)$ |
| SVR | $O\left(\frac{d}{k}\right) \cdot e^{O\left(\log^2 \frac{1}{\varepsilon}\right)}$ | $O\left(e^{\frac{d^2}{k\varepsilon^2}}\right)$ |

Table 1. Our sample complexity bounds.

Our bounds imply that for reaching a certain accuracy, our algorithms need *the same number of attributes* as their full information counterparts. In particular, when $k = \Omega(d)$ our bounds coincide with those of full information regression, up to constants (cf. Kakade et al. 2008).

We complement these upper bounds and prove that $\Omega(\frac{d}{\varepsilon^2})$ attributes are in fact necessary to learn an $\varepsilon$- accurate Ridge regressor. For Lasso regression, Cesa-Bianchi et al. (2010) proved that $\Omega(\frac{d}{\varepsilon})$ attributes are necessary, and asked what is the correct dependence on the problem dimension. Our bounds imply that the number of attributes necessary for regression learning grows linearly with the problem dimensions.

---

[1] The previous bounds are due to (Cesa-Bianchi et al., 2010). For SVR, the bound is obtained by additionally incorporating the methods of (Cesa-Bianchi et al., 2011).

[2] For SVR, the number of attributes viewed per example is a random variable whose expectation is $k$.

The algorithms themselves are very simple to implement, and run in linear time. As we show in later sections, these theoretical improvements are clearly visible in experiments on standard datasets.

### 1.1. Related work

The setting of learning with limited attribute observation (LAO) was first put forth in (Ben-David & Dichterman, 1998), who coined the term "learning with restricted focus of attention". Cesa-Biachi et al. (2010) were the first to discuss linear prediction in the LAO setting, and gave an efficient algorithm (as well as lower bounds) for linear regression, which is the primary focus of this paper.

## 2. Setting and Result Statement

### 2.1. Linear regression

In the linear regression problem, each instance is a pair $(\mathbf{x}, y)$ of an attributes vector $\mathbf{x} \in \mathbb{R}^d$ and a target variable $y \in \mathbb{R}$. We assume the standard framework of statistical learning (Haussler, 1992), in which the pairs $(\mathbf{x}, y)$ follow a joint probability distribution $\mathcal{D}$ over $\mathbb{R}^d \times \mathbb{R}$. The goal of the learner is to find a vector $\mathbf{w}$ for which the linear rule $\hat{y} \leftarrow \mathbf{w}^\top \mathbf{x}$ provides a good prediction of the target $y$. To measure the performance of the prediction, we use a convex loss function $\ell(\hat{y}, y) : \mathbb{R}^2 \to \mathbb{R}$. The most common choice is the squared loss $\ell(\hat{y}, y) = \frac{1}{2}(\hat{y}-y)^2$, which stands for the popular least-squares regression. Hence, in terms of the distribution $\mathcal{D}$, the learner would like to find a regressor $\mathbf{w} \in \mathbb{R}^d$ with low expected loss, defined as

$$L_\mathcal{D}(\mathbf{w}) = \mathbf{E}_{(\mathbf{x}, y) \sim \mathcal{D}}[\ell(\mathbf{w}^\top \mathbf{x}, y)] \,. \quad (1)$$

The standard paradigm for learning such regressor is seeking a vector $\mathbf{w} \in \mathbb{R}^d$ that minimizes a trade-off between the expected loss and an additional regularization term, which is usually a norm of $\mathbf{w}$. An equivalent form of this optimization problem is obtained by replacing the regularization term with a proper constraint, giving rise to the problem

$$\min_{\mathbf{w} \in \mathbb{R}^d} L_\mathcal{D}(\mathbf{w}) \quad \text{s.t.} \quad \|\mathbf{w}\|_p \leqslant B \,, \quad (2)$$

where $B > 0$ is a regularization parameter and $p \geqslant 1$. The main variants of regression differ on the type of $\ell_p$ norm constraint as well as the loss functions in the above definition:

- **Ridge regression:** $p = 2$ and squared loss,
$$\ell(\hat{y}, y) = \tfrac{1}{2}(\hat{y} - y)^2 \,.$$

- **Lasso regression:** $p = 1$ and squared loss.

- **Support-vector regression:** $p = 2$ and the $\delta$-insensitive absolute loss (Vapnik, 1995),
$$\ell(\hat{y}, y) = |\hat{y} - y|_\delta := \max\{0, |\hat{y} - y| - \delta\} \,.$$

Since the distribution $\mathcal{D}$ is unknown, we learn by relying on a training set $S = \{(\mathbf{x}_t, y_t)\}_{t=1}^m$ of examples, that are assumed to be sampled independently from $\mathcal{D}$. We use the notation $\ell_t(\mathbf{w}) := \ell(\mathbf{w}^\top \mathbf{x}_t, y_t)$ to refer to the loss function induced by the instance $(\mathbf{x}_t, y_t)$.

We distinguish between two learning scenarios. In the **full information** setup, the learner has unrestricted access to the entire data set. In the **limited attribute observation (LAO)** setting, for any given example pair $(\mathbf{x}, y)$, the learner can observe $y$, but only $k$ attributes of $\mathbf{x}$ (where $k \geqslant 1$ is a parameter of the problem). The learner can *actively choose* which attributes to observe.

### 2.2. Limitations on LAO regression

Cesa-Biachi et al. (2010) proved the following sample complexity lower bound on any LAO Lasso regression algorithm.

**Theorem 2.1.** *Let $0 < \varepsilon < \frac{1}{16}$, $k \geqslant 1$ and $d > 4k$. For any regression algorithm accessing at most $k$ attributes per training example, there exist a distribution $\mathcal{D}$ over $\{\mathbf{x} : \|\mathbf{x}\|_\infty \leqslant 1\} \times \{\pm 1\}$ and a regressor $\mathbf{w}_\star$ with $\|\mathbf{w}_\star\|_1 \leqslant 1$ such that the algorithm must see (in expectation) at least $\Omega(\frac{d}{k\varepsilon})$ examples in order to learn a linear regressor $\mathbf{w}$ with $L_\mathcal{D}(\mathbf{w}) - L_\mathcal{D}(\mathbf{w}_\star) < \varepsilon$.*

We complement this lower bound, by providing a stronger lower bound on the sample complexity of any Ridge regression algorithm, using information-theoretic arguments.

**Theorem 2.2.** *Let $\varepsilon = \Omega(1/\sqrt{d})$. For any regression algorithm accessing at most $k$ attributes per training example, there exist a distribution $\mathcal{D}$ over $\{\mathbf{x} : \|\mathbf{x}\|_2 \leqslant 1\} \times \{\pm 1\}$ and a regressor $\mathbf{w}_\star$ with $\|\mathbf{w}_\star\|_2 \leqslant 1$ such that the algorithm must see (in expectation) at least $\Omega(\frac{d}{k\varepsilon^2})$ examples in order to learn a linear regressor $\mathbf{w}$, $\|\mathbf{w}\|_2 \leqslant 1$ with $L_\mathcal{D}(\mathbf{w}) - L_\mathcal{D}(\mathbf{w}_\star) \leqslant \varepsilon$.*

Our algorithm for LAO Ridge regression (see section 3) imply this lower bound to be tight up to constants.

Note, however, that the bound applies only to a particular regime of the problem parameters [3].

### 2.3. Our algorithmic results

We give efficient regression algorithms that attain the following risk bounds. For our Ridge regression algorithm, we prove the risk bound

$$\mathbf{E}\left[L_{\mathcal{D}}(\bar{\mathbf{w}})\right] \leqslant \min_{\|\mathbf{w}\|_2 \leqslant B} L_{\mathcal{D}}(\mathbf{w}) + O\left(B^2 \sqrt{\frac{d}{km}}\right),$$

while for our Lasso regression algorithm we establish the bound

$$\mathbf{E}\left[L_{\mathcal{D}}(\bar{\mathbf{w}})\right] \leqslant \min_{\|\mathbf{w}\|_1 \leqslant B} L_{\mathcal{D}}(\mathbf{w}) + O\left(B^2 \sqrt{\frac{d \log d}{km}}\right).$$

Here we use $\bar{\mathbf{w}}$ to denote the output of each algorithm on a training set of $m$ examples, and the expectations are taken with respect to the randomization of the algorithms. For Support-vector regression we obtain a risk bound that depends on the desired accuracy $\varepsilon$. Our bound implies that

$$m = O\left(\frac{d}{k}\right) \cdot \exp\left(O\left(\log^2 \frac{B}{\varepsilon}\right)\right).$$

examples are needed (in expectation) for obtaining an $\varepsilon$-accurate regressor.

## 3. Algorithms for LAO least-squares regression

In this section we present and analyze our algorithms for Ridge and Lasso regression in the LAO setting. The loss function under consideration here is the squared loss, that is, $\ell_t(\mathbf{w}) = \frac{1}{2}(\mathbf{w}^\top \mathbf{x}_t - y_t)^2$. For convenience, we show algorithms that use $k+1$ attributes of each instance, for $k \geqslant 1$ [4].

Our algorithms are iterative and maintain a regressor $\mathbf{w}_t$ along the iterations. The update of the regressor at iteration $t$ is based on gradient information, and specifically on $\mathbf{g}_t := \nabla \ell_t(\mathbf{w}_t)$ that equals $(\mathbf{w}_t^\top \mathbf{x}_t - y_t) \cdot \mathbf{x}_t$ for the squared loss. In the LAO setting, however, we do not have the access to this information, thus we build upon unbiased estimators of the gradients.

---

[3] Indeed, there are (full-information) algorithms that are known to converge in $O(1/\varepsilon)$ rate – see e.g. (Hazan et al., 2007).

[4] We note that by our approach it is impossible to learn using a *single* attribute of each example (i.e., with $k=0$), and we are not aware of any algorithm that is able to do so. See (Cesa-Bianchi et al., 2011) for a related impossibility result.

---

**Algorithm 1** AERR
Parameters: $B, \eta > 0$
**Input:** training set $S = \{(\mathbf{x}_t, y_t)\}_{t \in [m]}$ and $k > 0$
**Output:** regressor $\bar{\mathbf{w}}$ with $\|\bar{\mathbf{w}}\|_2 \leqslant B$
1: Initialize $\mathbf{w}_1 \neq \mathbf{0}, \|\mathbf{w}_1\|_2 \leqslant B$ arbitrarily
2: **for** $t = 1$ **to** $m$ **do**
3:   **for** $r = 1$ **to** $k$ **do**
4:     Pick $i_{t,r} \in [d]$ uniformly and observe $\mathbf{x}_t[i_{t,r}]$
5:     $\tilde{\mathbf{x}}_{t,r} \leftarrow d\,\mathbf{x}_t[i_{t,r}] \cdot \mathbf{e}_{i_{t,r}}$
6:   **end for**
7:   $\tilde{\mathbf{x}}_t \leftarrow \frac{1}{k}\sum_{r=1}^k \tilde{\mathbf{x}}_{t,r}$
8:   Choose $j_t \in [d]$ with probability $\mathbf{w}_t[j]^2/\|\mathbf{w}_t\|_2^2$, and observe $\mathbf{x}_t[j_t]$
9:   $\tilde{\phi}_t \leftarrow \|\mathbf{w}_t\|_2^2\, \mathbf{x}_t[j_t]/\mathbf{w}_t[j_t] - y_t$
10:   $\tilde{\mathbf{g}}_t \leftarrow \tilde{\phi}_t \cdot \tilde{\mathbf{x}}_t$
11:   $\mathbf{v}_t \leftarrow \mathbf{w}_t - \eta \tilde{\mathbf{g}}_t$
12:   $\mathbf{w}_{t+1} \leftarrow \mathbf{v}_t \cdot B/\max\{\|\mathbf{v}_t\|_2, B\}$
13: **end for**
14: $\bar{\mathbf{w}} \leftarrow \frac{1}{m}\sum_{t=1}^m \mathbf{w}_t$

### 3.1. Ridge regression

Recall that in Ridge regression, we are interested in the linear regressor that is the solution to the optimization problem (2) with $p = 2$, given explicitly as

$$\min_{\mathbf{w} \in \mathbb{R}^d} L_{\mathcal{D}}(\mathbf{w}) \quad \text{s.t.} \quad \|\mathbf{w}\|_2 \leqslant B, \qquad (3)$$

Our algorithm for the LAO setting is based on a randomized Online Gradient Descent (OGD) strategy (Zinkevich, 2003). More specifically, at each iteration $t$ we use a randomized estimator $\tilde{\mathbf{g}}_t$ of the gradient $\mathbf{g}_t$ to update the regressor $\mathbf{w}_t$ via an additive rule. Our gradient estimators make use of an importance-sampling method inspired by (Clarkson et al., 2010).

The pseudo-code of our Attribute Efficient Ridge Regression (AERR) algorithm is given in Algorithm 1. In the following theorem, we show that the regressor learned by our algorithm is competitive with the optimal linear regressor having 2-norm bounded by $B$.

**Theorem 3.1.** *Assume the distribution $\mathcal{D}$ is such that $\|\mathbf{x}\|_2 \leqslant 1$ and $|y| \leqslant B$ with probability 1. Let $\bar{\mathbf{w}}$ be the output of AERR, when run with $\eta = \sqrt{k/2dm}$. Then, $\|\bar{\mathbf{w}}\|_2 \leqslant B$ and for any $\mathbf{w}_\star \in \mathbb{R}^d$ with $\|\mathbf{w}_\star\|_2 \leqslant B$,*

$$\mathbf{E}\left[L_{\mathcal{D}}(\bar{\mathbf{w}})\right] \leqslant L_{\mathcal{D}}(\mathbf{w}_\star) + 4B^2 \sqrt{\frac{2d}{km}}.$$

#### 3.1.1. Analysis

Theorem 3.1 is a consequence of the following two lemmas. The first lemma is obtained as a result of a standard regret bound for the OGD algorithm (see Zinkevich 2003), applied to the vectors $\tilde{\mathbf{g}}_1, \ldots, \tilde{\mathbf{g}}_m$.

**Lemma 3.2.** *For any $\|\mathbf{w}_\star\|_2 \leqslant B$ we have*

$$\sum_{t=1}^m \tilde{\mathbf{g}}_t^\top (\mathbf{w}_t - \mathbf{w}_\star) \leqslant \frac{2B^2}{\eta} + \frac{\eta}{2} \sum_{t=1}^m \|\tilde{\mathbf{g}}_t\|_2^2 \ . \qquad (4)$$

The second lemma shows that the vector $\tilde{\mathbf{g}}_t$ is an unbiased estimator of the gradient $\mathbf{g}_t := \nabla \ell_t(\mathbf{w}_t)$ at iteration $t$, and establishes a "variance" bound for this estimator. To simplify notations, here and in the rest of the paper we use $\mathbf{E}_t[\cdot]$ to denote the conditional expectation with respect to all randomness up to time $t$.

**Lemma 3.3.** *The vector $\tilde{\mathbf{g}}_t$ is an unbiased estimator of the gradient $\mathbf{g}_t := \nabla \ell_t(\mathbf{w}_t)$, that is $\mathbf{E}_t[\tilde{\mathbf{g}}_t] = \mathbf{g}_t$. In addition, for all $t$ we have $\mathbf{E}_t[\|\tilde{\mathbf{g}}_t\|_2^2] \leqslant 8B^2 d/k$.*

For a proof of the lemma, see (Hazan & Koren, 2011). We now turn to prove Theorem 3.1.

*Proof (of Theorem 3.1).* First note that as $\|\mathbf{w}_t\|_2 \leqslant B$, we clearly have $\|\bar{\mathbf{w}}\|_2 \leqslant B$. Taking the expectation of (4) with respect to the randomization of the algorithm, and letting $G^2 := \max_t \mathbf{E}_t[\|\tilde{\mathbf{g}}_t\|_2^2]$, we obtain

$$\mathbf{E}\left[\sum_{t=1}^m \mathbf{g}_t^\top (\mathbf{w}_t - \mathbf{w}_\star)\right] \leqslant \frac{2B^2}{\eta} + \frac{\eta}{2} G^2 m \ .$$

On the other hand, the convexity of $\ell_t$ gives $\ell_t(\mathbf{w}_t) - \ell_t(\mathbf{w}_\star) \leqslant \mathbf{g}_t^\top (\mathbf{w}_t - \mathbf{w}_\star)$. Together with the above this implies that for $\eta = 2B/G\sqrt{m}$,

$$\mathbf{E}\left[\frac{1}{m}\sum_{t=1}^m \ell_t(\mathbf{w}_t)\right] \leqslant \frac{1}{m}\sum_{t=1}^m \ell_t(\mathbf{w}_\star) + 2\frac{BG}{\sqrt{m}} \ .$$

Taking the expectation of both sides with respect to the random choice of the training set, and using $G \leqslant 2B\sqrt{2d/k}$ (according to Lemma 3.3), we get

$$\mathbf{E}\left[\frac{1}{m}\sum_{t=1}^m L_\mathcal{D}(\mathbf{w}_t)\right] \leqslant L_\mathcal{D}(\mathbf{w}_\star) + 4B^2 \sqrt{\frac{2d}{km}} \ .$$

Finally, recalling the convexity of $L_\mathcal{D}$ and using Jensen's inequality, the Theorem follows. $\square$

### 3.2. Lasso regression

We now turn to describe our algorithm for Lasso regression in the LAO setting, in which we would like to solve the problem

$$\min_{\mathbf{w} \in \mathbb{R}^d} L_\mathcal{D}(\mathbf{w}) \quad \text{s.t.} \quad \|\mathbf{w}\|_1 \leqslant B \ . \qquad (5)$$

The algorithm we provide for this problem is based on a stochastic variant of the EG algorithm (Kivinen & Warmuth, 1997), that employs *multiplicative* updates based on an estimation of the gradients $\nabla \ell_t$. The multiplicative nature of the algorithm, however, makes it highly sensitive to the magnitude of the updates. To make the updates more robust, we "clip" the entries of the gradient estimator so as to prevent them from getting too large. Formally, this is accomplished via the following "clip" operation:

$$\mathrm{clip}(x,c) := \max\{\min\{x,c\}, -c\}$$

for $x \in \mathbb{R}$ and $c > 0$. This clipping has an even stronger effect in the more general setting we consider in Section 4.

We give our Attribute Efficient Lasso Regression (AELR) algorithm in Algorithm 2, and establish a corresponding risk bound in the following theorem.

**Algorithm 2** AELR

Parameters: $B, \eta > 0$

**Input:** training set $S = \{(\mathbf{x}_t, y_t)\}_{t \in [m]}$ and $k > 0$
**Output:** regressor $\bar{\mathbf{w}}$ with $\|\bar{\mathbf{w}}\|_1 \leqslant B$
1: Initialize $\mathbf{z}_1^+ \leftarrow \mathbf{1}_d$ , $\mathbf{z}_1^- \leftarrow \mathbf{1}_d$
2: **for** $t = 1$ to $m$ **do**
3: $\quad \mathbf{w}_t \leftarrow (\mathbf{z}_t^+ - \mathbf{z}_t^-) \cdot B/(\|\mathbf{z}_t^+\|_1 + \|\mathbf{z}_t^-\|_1)$
4: $\quad$ **for** $r = 1$ **to** $k$ **do**
5: $\quad\quad$ Pick $i_{t,r} \in [d]$ uniformly and observe $\mathbf{x}_t[i_{t,r}]$
6: $\quad\quad \tilde{\mathbf{x}}_{t,r} \leftarrow d\,\mathbf{x}_t[i_{t,r}] \cdot \mathbf{e}_{i_{t,r}}$
7: $\quad$ **end for**
8: $\quad \tilde{\mathbf{x}}_t \leftarrow \frac{1}{k}\sum_{r=1}^k \tilde{\mathbf{x}}_{t,r}$
9: $\quad$ Choose $j_t \in [d]$ with probability $|\mathbf{w}[j]|/\|\mathbf{w}\|_1$, and observe $\mathbf{x}_t[j_t]$
10: $\quad \tilde{\phi}_t \leftarrow \|\mathbf{w}_t\|_1 \,\mathrm{sign}(\mathbf{w}_t[j_t])\,\mathbf{x}_t[j_t] - y_t$
11: $\quad \tilde{\mathbf{g}}_t \leftarrow \tilde{\phi}_t \cdot \tilde{\mathbf{x}}_t$
12: $\quad$ **for** $i = 1$ to $d$ **do**
13: $\quad\quad \bar{\mathbf{g}}_t[i] \leftarrow \mathrm{clip}(\tilde{\mathbf{g}}_t[i], 1/\eta)$
14: $\quad\quad \mathbf{z}_{t+1}^+[i] \leftarrow \mathbf{z}_t^+[i] \cdot \exp(-\eta\,\bar{\mathbf{g}}_t[i])$
15: $\quad\quad \mathbf{z}_{t+1}^-[i] \leftarrow \mathbf{z}_t^-[i] \cdot \exp(+\eta\,\bar{\mathbf{g}}_t[i])$
16: $\quad$ **end for**
17: **end for**
18: $\bar{\mathbf{w}} \leftarrow \frac{1}{m}\sum_{t=1}^m \mathbf{w}_t$

**Theorem 3.4.** *Assume the distribution $\mathcal{D}$ is such that $\|\mathbf{x}\|_\infty \leqslant 1$ and $|y| \leqslant B$ with probability 1. Let $\bar{\mathbf{w}}$ be the output of AELR, when run with $\eta = \frac{1}{4B^2}\sqrt{\frac{2k \log 2d}{5md}}$, Then, $\|\bar{\mathbf{w}}\|_1 \leqslant B$ and for any $\mathbf{w}_\star \in \mathbb{R}^d$ with $\|\mathbf{w}_\star\|_1 \leqslant B$ we have*

$$\mathbf{E}\left[L_\mathcal{D}(\bar{\mathbf{w}})\right] \leqslant L_\mathcal{D}(\mathbf{w}_\star) + 4B^2 \sqrt{\frac{10d \log 2d}{km}} \ ,$$

*provided that $m \geqslant \log 2d$.*

### 3.2.1. ANALYSIS

In the rest of the section, for a vector $\mathbf{v}$ we let $\mathbf{v}^2$ denote the vector for which $\mathbf{v}^2[i] = (\mathbf{v}[i])^2$ for all $i$.

In order to prove Theorem 3.4, we first consider the augmented vectors $\mathbf{z}'_t := (\mathbf{z}_t^+, \mathbf{z}_t^-) \in \mathbb{R}^{2d}$ and $\bar{\mathbf{g}}'_t := (\bar{\mathbf{g}}_t, -\bar{\mathbf{g}}_t) \in \mathbb{R}^{2d}$, and let $\mathbf{p}_t := \mathbf{z}'_t/\|\mathbf{z}'_t\|_1$. For these vectors, we have the following.

**Lemma 3.5.**
$$\sum_{t=1}^m \mathbf{p}_t^\top \bar{\mathbf{g}}'_t \leq \min_{i \in [2d]} \sum_{t=1}^m \bar{\mathbf{g}}'_t[i] + \frac{\log 2d}{\eta} + \eta \sum_{t=1}^m \mathbf{p}_t^\top (\bar{\mathbf{g}}'_t)^2$$

The lemma is a consequence of a second-order regret bound for the Multiplicative-Weights algorithm, essentially due to (Clarkson et al., 2010). By means of this lemma, we establish a risk bound with respect to the "clipped" linear functions $\bar{\mathbf{g}}_t^\top \mathbf{w}$.

**Lemma 3.6.** *Assume that $\|\mathbf{E}_t[\tilde{\mathbf{g}}_t^2]\|_\infty \leq G^2$ for all $t$, for some $G > 0$. Then, for any $\|\mathbf{w}_\star\|_1 \leq B$, we have*
$$\mathbf{E}\left[\sum_{t=1}^m \bar{\mathbf{g}}_t^\top \mathbf{w}_t\right] \leq \mathbf{E}\left[\sum_{t=1}^m \bar{\mathbf{g}}_t^\top \mathbf{w}_\star\right] + B\left(\frac{\log 2d}{\eta} + \eta G^2 m\right)$$

Our next step is to relate the risk generated by the linear functions $\tilde{\mathbf{g}}_t^\top \mathbf{w}$, to that generated by the "clipped" functions $\bar{\mathbf{g}}_t^\top \mathbf{w}$.

**Lemma 3.7.** *Assume that $\|\mathbf{E}_t[\tilde{\mathbf{g}}_t^2]\|_\infty \leq G^2$ for all $t$, for some $G > 0$. Then, for $0 < \eta \leq 1/2G$ we have*
$$\mathbf{E}\left[\sum_{t=1}^m \tilde{\mathbf{g}}_t^\top \mathbf{w}_t\right] \leq \mathbf{E}\left[\sum_{t=1}^m \bar{\mathbf{g}}_t^\top \mathbf{w}_t\right] + 4B\eta G^2 m \ .$$

The final component of the proof is a "variance" bound, similar to that of Lemma 3.3.

**Lemma 3.8.** *The vector $\tilde{\mathbf{g}}_t$ is an unbiased estimator of the gradient $\mathbf{g}_t := \nabla \ell_t(\mathbf{w}_t)$, that is $\mathbf{E}_t[\tilde{\mathbf{g}}_t] = \mathbf{g}_t$. In addition, for all $t$ we have $\|\mathbf{E}_t[\tilde{\mathbf{g}}_t]^2\|_\infty \leq 8B^2 d/k$.*

For the complete proofs, refer to (Hazan & Koren, 2011). We are now ready to prove Theorem 3.4.

*Proof (of Theorem 3.4).* Since $\|\mathbf{w}_t\|_1 \leq B$ for all $t$, we obtain $\|\bar{\mathbf{w}}\|_2 \leq B$. Next, note that as $\mathbf{E}_t[\tilde{\mathbf{g}}_t] = \mathbf{g}_t$, we have $\mathbf{E}[\sum_{t=1}^m \tilde{\mathbf{g}}_t^\top \mathbf{w}_t] = \mathbf{E}[\sum_{t=1}^m \mathbf{g}_t^\top \mathbf{w}_t]$. Putting Lemmas 3.6 and 3.7 together, we get for $\eta \leq 1/2G$ that
$$\mathbf{E}\left[\sum_{t=1}^T \mathbf{g}_t^\top (\mathbf{w}_t - \mathbf{w}_\star)\right] \leq B\left(\frac{\log 2d}{\eta} + 5\eta G^2 m\right) \ .$$

Proceeding as in the proof of Theorem 3.1, and choosing $\eta = \frac{1}{G}\sqrt{\frac{\log 2d}{5m}}$, we obtain the bound
$$\mathbf{E}\left[L_\mathcal{D}(\bar{\mathbf{w}})\right] \leq L_\mathcal{D}(\mathbf{w}_\star) + 2BG\sqrt{\frac{5 \log 2d}{m}} \ .$$

Note that for this choice of $\eta$ we indeed have $\eta \leq 1/2G$, as we originally assumed that $m \geq \log 2d$. Finally, putting $G = 2B\sqrt{2d/k}$ as implied by Lemma 3.8, we obtain the bound in the statement of the theorem. □

## 4. Support-vector regression

In this section we show how our approach can be extended to deal with loss functions other than the squared loss, of the form

$$\ell(\mathbf{w}^\top \mathbf{x}, y) = f(\mathbf{w}^\top \mathbf{x} - y) \ , \qquad (6)$$

(with $f$ real and convex) and most importantly, with the $\delta$-insensitive absolute loss function of SVR, for which $f(x) = |x|_\delta := \max\{|x| - \delta, 0\}$ for some fixed $0 \leq \delta \leq B$ (recall that in our results we assume the labels $y_t$ have $|y_t| \leq B$). For concreteness, we consider only the 2-norm variant of the problem (as in the standard formulation of SVR)—the results we obtain can be easily adjusted to the 1-norm setting. We overload notation, and keep using the shorthand $\ell_t(\mathbf{w}) := \ell(\mathbf{w}^\top \mathbf{x}_t, y_t)$ for referring the loss function induced by the instance $(\mathbf{x}_t, y_t)$.

It should be highlighted that our techniques can be adapted to deal with many other common loss functions, including "classification" losses (i.e., of the form $\ell(\mathbf{w}^\top \mathbf{x}, y) = f(y \cdot \mathbf{w}^\top \mathbf{x})$). Due to its importance and popularity, we chose to describe our method in the context of SVR.

Unfortunately, there are strong indications that SVR learning (more generally, learning with non-smooth loss function) in the LAO setting is impossible via our approach of unbiased gradient estimations (see Cesa-Bianchi et al. 2011 and the references therein). For that reason, we make two modifications to the learning setting: first, we shall henceforth relax the budget constraint to allow $k$ observed attributes per instance *in expectation*; and second, we shall aim for *biased* gradient estimators, instead of unbiased as before.

To obtain such biased estimators, we uniformly $\varepsilon$-approximate the function $f$ by an *analytic function* $f_\varepsilon$ and learn with the approximate loss function $\ell_t^\varepsilon(\mathbf{w}) = f_\varepsilon(\mathbf{w}^\top \mathbf{x}_t - y_t)$ instead. Clearly, any $\varepsilon$-suboptimal regressor of the approximate problem is an $2\varepsilon$-suboptimal regressor of the original problem. For learning the approximate problem we use a novel technique, inspired by (Cesa-Bianchi et al., 2011), for estimating gradients of analytic loss functions. Our estimators for $\nabla \ell_t^\varepsilon$ can then be viewed as biased estimators of $\nabla \ell_t$ (we note, however, that the resulting bias might be quite large).

**Procedure 3** GenEst

Parameters: $\{a_n\}_{n=0}^{\infty}$ — Taylor coefficients of $f'$

**Input:** regressor $\mathbf{w}$, instance $(\mathbf{x}, y)$
**Output:** $\hat{\phi}$ with $\mathbf{E}[\hat{\phi}] = f'(\mathbf{w}^\top \mathbf{x} - y)$

1: Let $N = \lceil 4B^2 \rceil$.
2: Choose $n \geq 0$ with probability $\Pr[n] = (\frac{1}{2})^{n+1}$
3: **if** $n \leq 2 \log_2 N$ **then**
4:   **for** $r = 1, \ldots, n$ **do**
5:     Choose $j \in [d]$ with probability $\mathbf{w}[j]^2 / \|\mathbf{w}\|_2^2$, and observe $\mathbf{x}[j]$
6:     $\tilde{\theta}_r \leftarrow \|\mathbf{w}\|_2^2 \mathbf{x}[j] / \mathbf{w}[j] - y$
7:   **end for**
8: **else**
9:   **for** $r = 1, \ldots, n$ **do**
10:    Choose $j_1, \ldots, j_N \in [d]$ w.p. $\mathbf{w}[j]^2/\|\mathbf{w}\|_2^2$, (independently), and observe $\mathbf{x}[j_1], \ldots, \mathbf{x}[j_N]$
11:    $\tilde{\theta}_r \leftarrow \frac{1}{N} \sum_{s=1}^{N} \|\mathbf{w}\|_2^2 \mathbf{x}[j_s] / \mathbf{w}[j_s] - y$
12:  **end for**
13: **end if**
14: $\hat{\phi} \leftarrow 2^{n+1} a_n \cdot \tilde{\theta}_1 \tilde{\theta}_2 \cdots \tilde{\theta}_n$

### 4.1. Estimators for analytic loss functions

Let $f : \mathbb{R} \to \mathbb{R}$ be a real, analytic function (on the entire real line). The derivative $f'$ is thus also analytic and can be expressed as $f'(x) = \sum_{n=0}^{\infty} a_n x^n$ where $\{a_n\}$ are the Taylor expansion coefficients of $f'$.

In Procedure 3 we give an unbiased estimator of $f'(\mathbf{w}^\top \mathbf{x} - y)$ in the LAO setting, defined in terms of the coefficients $\{a_n\}$ of $f'$. For this estimator, we have the following (proof is omitted).

**Lemma 4.1.** *The estimator $\hat{\phi}$ is an unbiased estimator of $f'(\mathbf{w}^\top \mathbf{x} - y)$. Also, assuming $\|\mathbf{x}\|_2 \leq 1$, $\|\mathbf{w}\|_2 \leq B$ and $|y| \leq B$, the second-moment $\mathbf{E}[\hat{\phi}^2]$ is upper bounded by $\exp(O(\log^2 B))$, provided that the Taylor series of $f'(x)$ converges absolutely for $|x| \leq 1$. Finally, the expected number of attributes of $\mathbf{x}$ used by this estimator is no more than 3.*

### 4.2. Approximating SVR

In order to approximate the $\delta$-insensitive absolute loss function, we define

$$f_\varepsilon(x) = \frac{\varepsilon}{2} \rho\left(\frac{x-\delta}{\varepsilon}\right) + \frac{\varepsilon}{2} \rho\left(\frac{x+\delta}{\varepsilon}\right) - \delta$$

where $\rho$ is expressed in terms of the error function $\mathrm{erf}$,

$$\rho(x) = x \, \mathrm{erf}(x) + \frac{1}{\sqrt{\pi}} e^{-x^2},$$

and consider the approximate loss functions $\ell_t^\varepsilon(\mathbf{w}) = f_\varepsilon(\mathbf{w}^\top \mathbf{x}_t - y_t)$. Indeed, we have the following.

**Algorithm 4** AESVR

Parameters: $B, \delta, \eta > 0$ and accuracy $\varepsilon > 0$
**Input:** training set $S = \{(\mathbf{x}_t, y_t)\}_{t \in [m]}$ and $k > 0$
**Output:** regressor $\bar{\mathbf{w}}$ with $\|\bar{\mathbf{w}}\|_2 \leq B$

1: Let $a_{2n} = 0$ for $n \geq 0$, and

$$a_{2n+1} = \frac{2}{\sqrt{\pi}} \cdot \frac{(-1)^n}{n!(2n+1)}, \qquad n \geq 0 \quad (7)$$

2: Execute algorithm 1 with lines 8–9 replaced by:
  $\mathbf{x}'_t \leftarrow \mathbf{x}_t / \varepsilon$
  $y_t^+ \leftarrow (y_t + \delta)/\varepsilon, \quad y_t^- \leftarrow (y_t - \delta)/\varepsilon$
  $\tilde{\phi}_t \leftarrow \frac{1}{2}[\mathtt{GenEst}(\mathbf{w}_t, \mathbf{x}'_t, y_t^+) + \mathtt{GenEst}(\mathbf{w}_t, \mathbf{x}'_t, y_t^-)]$
3: Return the output $\bar{\mathbf{w}}$ of the algorithm

**Claim 4.2.** *For any $\varepsilon > 0$, $f_\varepsilon$ is convex, analytic on the entire real line and*

$$\sup_{x \in \mathbb{R}} |f_\varepsilon(x) - |x|_\delta| \leq \varepsilon \,.$$

The claim follows easily from the identity $|x|_\delta = \frac{1}{2}|x - \delta| + \frac{1}{2}|x + \delta| - \delta$. In addition, for using Procedure 3 we need the following simple observation, that follows immediately from the series expansion of $\mathrm{erf}(x)$.

**Claim 4.3.** *$\rho'(x) = \sum_{n=0}^{\infty} a_{2n+1} x^{2n+1}$, with the coefficients $\{a_{2n+1}\}_{n \geq 0}$ given in (7).*

We now give the main result of this section, which is a sample complexity bound for the Attribute Efficient SVR (AESVR) algorithm, given in Algorithm 4.

**Theorem 4.4.** *Assume the distribution $\mathcal{D}$ is such that $\|\mathbf{x}\|_2 \leq 1$ and $|y| \leq B$ with probability 1. Then, for any $\mathbf{w}_\star \in \mathbb{R}^d$ with $\|\mathbf{w}_\star\|_2 \leq B$, we have $\mathbf{E}[L_\mathcal{D}(\bar{\mathbf{w}})] \leq L_\mathcal{D}(\mathbf{w}_\star) + \varepsilon$ where $\bar{\mathbf{w}}$ is the output of AESVR (with $\eta$ properly tuned) on a training set of size*

$$m = O\left(\frac{d}{k}\right) \cdot \exp\left(O\left(\log^2 \frac{B}{\varepsilon}\right)\right). \quad (8)$$

*The algorithm queries at most $k+6$ attributes of each instance in expectation.*

*Proof.* First, note that for the approximate loss functions $\ell_t^\varepsilon$ we have

$$\nabla \ell_t^\varepsilon(\mathbf{w}_t) = \tfrac{1}{2} \left[ \rho'(\mathbf{w}_t^\top \mathbf{x}'_t - y_t^+) + \rho'(\mathbf{w}_t^\top \mathbf{x}'_t - y_t^-) \right] \cdot \mathbf{x}_t \,.$$

Hence, Lemma 4.1 and Claim 4.3 above imply that $\tilde{\mathbf{g}}_t$ in Algorithm 4 is an unbiased estimator of $\nabla \ell_t^\varepsilon(\mathbf{w}_t)$. Furthermore, since $\|\mathbf{x}'_t\|_2 \leq \frac{1}{\varepsilon}$ and $|y_t^\pm| \leq 2\frac{B}{\varepsilon}$, according to the same lemma we have $\mathbf{E}_t[\tilde{\phi}_t^2] = \exp(O(\log^2 \frac{B}{\varepsilon}))$. Repeating the proof of Lemma 3.3,

we then have

$$\mathbf{E}_t[\|\tilde{\mathbf{g}}_t\|_2^2] = \mathbf{E}_t[\tilde{\phi}_t^2] \cdot \mathbf{E}_t[\|\tilde{\mathbf{x}}_t\|_2^2] = \exp(O(\log^2 \tfrac{B}{\varepsilon})) \cdot \frac{d}{k} \ .$$

Replacing $G^2$ in the proof of theorem 3.1 with the above bound, we get for the output of Algorithm 4,

$$\mathbf{E}\left[L_{\mathcal{D}}(\bar{\mathbf{w}})\right] \leq L_{\mathcal{D}}(\mathbf{w}_\star) + \exp(O(\log^2 \tfrac{B}{\varepsilon}))\sqrt{\frac{d}{km}},$$

which imply that for obtaining an $\varepsilon$-accurate regressor $\bar{\mathbf{w}}$ of the approximate problem, it is enough to take $m$ as given in (8). However, claim 4.2 now gives that $\bar{\mathbf{w}}$ itself is an $2\varepsilon$-accurate regressor of the original problem, and the proof is complete. □

## 5. Experiments

In this section we give experimental evidence that support our theoretical bounds, and demonstrate the superior performance of our algorithms compared to the state of the art. Naturally, we chose to compare our AERR and AELR algorithms [5] with the AER algorithm of (Cesa-Bianchi et al., 2010). We note that AER is in fact a hybrid algorithm that combines 1-norm and 2-norm regularizations, thus we use it for benchmarking in both the Ridge and Lasso settings.

We essentially repeated the experiments of (Cesa-Bianchi et al., 2010) and used the popular MNIST digit recognition dataset (LeCun et al., 1998). Each instance in this dataset is a $28 \times 28$ image of a handwritten digit $0-9$. We focused on the "3 vs. 5" task, on a subset of the dataset that consists of the "3" digits (labeled $-1$) and the "5" digits (labeled $+1$). We applied the regression algorithms to this task by regressing to the labels.

In all our experiments, we randomly split the data to training and test sets, and used 10-fold cross-validation for tuning the parameters of each algorithm. Then, we ran each algorithm on increasingly longer prefixes of the dataset and tracked the obtained squared-error on the test set. For faithfully comparing partial- and full-information algorithms, we also recorded the total *number of attributes* used by each algorithm.

In our first experiment, we executed AELR, AER and (offline) Lasso on the "3 vs. 5" task. We allowed both AELR and AER to use only $k = 4$ pixels of each training image, while giving Lasso unrestricted access to the entire set of attributes (total of 784) of each instance. The results, averaged over 10 runs on random train/test splits, are presented in Figure 1. Note that the $x$-axis represents the cumulative number of attributes used for training. The graph ends at roughly 48500 attributes, which is the total number of attributes allowed for the partial-information algorithms. Lasso, however, completes this budget after seeing merely 62 examples.

---
[5] The AESVR algorithm is presented mainly for theoretical considerations, and was not implemented in the experiments.

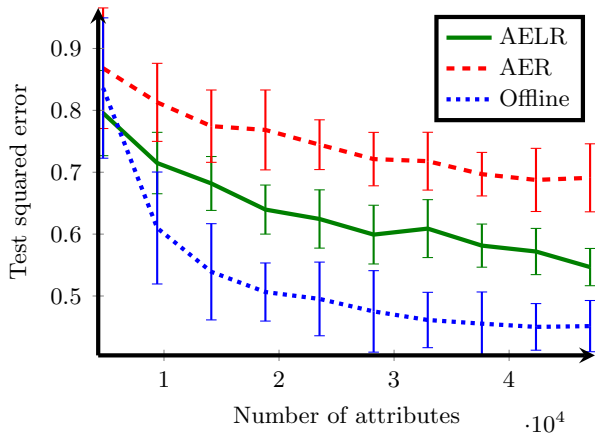

Figure 1. Test squared error of Lasso algorithms with $k = 4$, over increasing prefixes of the "3 vs. 5" dataset.

As we see from the results, AELR keeps its test error significantly lower than that of AER along the entire execution, almost bridging the gap with the full-information Lasso. Note that the latter has the clear advantage of being an offline algorithm, while both AELR and AER are online in nature. Indeed, when we compared AELR with an *online* Lasso solver, our algorithm obtained test error almost 10 times better.

In the second experiment, we evaluated AERR, AER and Ridge regression on the same task, but now allowing the partial-information algorithms to use as much as $k = 56$ pixels (which amounts to 2 rows) of each instance. The results of this experiment are given in Figure 2. We see that even if we allow the algorithms to view a considerable number of attributes, the gap between AERR and AER is large.

## 6. Conclusions and Open Questions

We have considered the fundamental problem of statistical regression analysis, and in particular Lasso and Ridge regression, in a setting where the observation upon each training instance is limited to a few attributes, and gave algorithms that improve over the state of the art by a leading order term with respect to the sample complexity. This resolves an open question of (Cesa-Bianchi et al., 2010). The algorithms are efficient, and give a clear experimental advantage in previously-considered benchmarks.

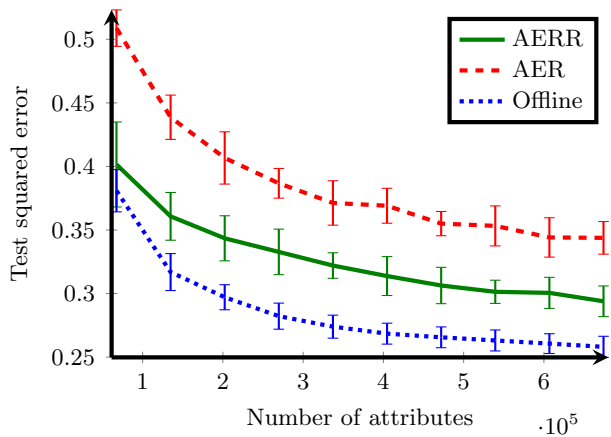

Figure 2. Test squared error of Ridge algorithms with $k = 56$, over increasing prefixes of the "3 vs. 5" dataset.

For the challenging case of regression with general convex loss functions, we describe exponential improvement in sample complexity, which apply in particular to support-vector regression.

It is interesting to resolve the sample complexity gap of $\frac{1}{\varepsilon}$ which still remains for Lasso regression, and to improve upon the pseudo-polynomial factor in $\varepsilon$ for support-vector regression. In addition, establishing analogous bounds for our algorithms that hold with high probability (other than in expectation) appears to be non-trivial, and is left for future work.

Another possible direction for future research is adapting our results to the setting of learning with (randomly) missing data, that was recently investigated—see e.g. (Rostamizadeh et al., 2011; Loh & Wainwright, 2011). The sample complexity bounds our algorithms obtain in this setting are slightly worse than those presented in the current paper, and it is interesting to check if one can do better.

## Acknowledgments

We thank Shai Shalev-Shwartz for several useful discussions, and the anonymous referees for their detailed comments.